%% file: neurips2025.tex
\documentclass{article}

\usepackage[dblblindworkshop, final]{neurips2025}
\usepackage[utf8]{inputenc} % allow utf-8 input
\usepackage[T1]{fontenc}    % use 8-bit T1 fonts
\usepackage{hyperref}       % hyperlinks
\usepackage{url}            % simple URL typesetting
\usepackage{minted}
\usepackage{booktabs}       % professional-quality tables
\usepackage{amsfonts}       % blackboard math symbols
\usepackage{nicefrac}       % compact symbols for 1/2, etc.
\usepackage{microtype}      % microtypography
\usepackage{xcolor}         % colors
\usepackage{graphicx}
\usepackage{pifont}     % For the check and cross marks
\usepackage[dvipsnames]{xcolor} % For custom colors
\usepackage{graphicx}
\usepackage{multirow}

\newcommand{\cmark}{\ding{51}}%
\newcommand{\xmark}{\ding{55}}%
\newcommand{\yes}{\textcolor{OliveGreen}{\cmark}}
\newcommand{\no}{\textcolor{BrickRed}{\xmark}}

\title{\textsc{Quran-MD}: A Fine-Grained Multilingual Multimodal Dataset of the Quran}

\author{%
  Muhammad Umar Salman \\
  MBZUAI\\
  Abu Dhabi, UAE\\
  \texttt{umar.salman@mbzuai.ac.ae} \\
  \And
  Mohammad Areeb Qazi \\
  MBZUAI\\
  Abu Dhabi, UAE\\
  \texttt{mohammad.qazi@mbzuai.ac.ae} \\
   \And
  Mohammed Talha Alam \\
  MBZUAI\\
  Abu Dhabi, UAE\\
  \texttt{mohammed.alam@mbzuai.ac.ae} \\
}

\begin{document}

\maketitle

\begin{abstract}
  We present \textsc{Quran-MD}, a comprehensive multimodal dataset of the Qur’an that integrates textual, linguistic, and audio dimensions at the verse and word levels. For each verse (ayah), the dataset provides its original Arabic text, English translation, and phonetic transliteration. To capture the rich oral tradition of Qur’anic recitation, we include verse-level audio from 32 distinct reciters, reflecting diverse recitation styles and dialectical nuances. At the word level, each token is paired with its corresponding Arabic script, English translation, transliteration, and an aligned audio recording, allowing fine-grained analysis of pronunciation, phonology, and semantic context. This dataset supports various applications, including natural language processing, speech recognition, text-to-speech synthesis, linguistic analysis, and digital Islamic studies. Bridging text and audio modalities across multiple reciters, this dataset provides a unique resource to advance computational approaches to Qur’anic recitation and study. Beyond enabling tasks such as ASR, tajweed detection, and Qur’anic TTS, it lays the foundation for multimodal embeddings, semantic retrieval, style transfer, and personalized tutoring systems that can support both research and community applications. The dataset is available at \url{https://huggingface.co/datasets/Buraaq/quran-audio-text-dataset}

\end{abstract}

\input{section/1_Introduction} 
\input{section/2_RelatedWorks} 
\input{section/3-4section}

\input{section/5_FutureWorks}
\input{section/6_Conslusion} 
\input{section/7_appendix}

\bibliographystyle{unsrt}
\bibliography{references}

\end{document}

%% file: section/1_Introduction.tex
\section{Introduction}
\label{sec:intro}

The Qur’an, the holy book of Islam, occupies a central role in the spiritual, intellectual, and cultural lives of nearly two billion Muslims worldwide. Revealed in Classical Arabic (Fusha), it is not only revered as sacred scripture but also preserved through its oral tradition of recitation. Mastery of recitation is guided by tajweed which is a set of phonological and prosodic rules that ensure the Qur’an is read with precision, beauty, and correctness. For Muslims, engaging with the Qur’an involves more than simply reading the text but is an act of devotion that requires both accurate pronunciation and a deep understanding of its meaning.

Despite its origins in Arabic, the Qur’an is recited and studied across every corner of the globe, including by millions of non-Arabic speakers from diverse linguistic and cultural backgrounds. This global engagement highlights the challenges faced by learners who must navigate not only the Arabic script but also its sounds, phonology, and interpretive meanings. While Islamic scholars, including Ulama (religious scholars), Fuqaha (jurists), and researchers, have made significant contributions to Qur’anic studies, the integration of modern computational approaches, particularly artificial intelligence, remains very limited compared to the rapid advances in AI. AI has the potential to illuminate linguistic patterns, aid recitation training, and support both Arabic and non-Arabic learners in mastering tajweed and understanding the Qur’an’s message.

Existing resources, however, often fall short of capturing the Qur’an’s full multimodal nature. Some datasets provide the Arabic text alone, others include translations or transliterations, and a few extend to audio recordings at either the verse or word level. Rarely are these modalities combined in a way that allows fine-grained analysis of both the written and spoken Qur’an. This fragmentation restricts their utility for advancing AI-driven research in areas such as natural language processing, speech recognition, and text-to-speech synthesis. Table \ref{tab:comparison_final} summarizes key features of prior Qur’anic datasets, where \textsc{Qur’an-MD} offers the most comprehensive multimodal coverage.

In this work, we curate and harmonize resources from three data sources to construct \textsc{Qur’an-MD}, a comprehensive multimodal dataset of the Qur’an. Our dataset unifies Arabic text, English translation, and phonetic transliteration at both the verse and word levels, while pairing each word with an aligned audio recording of its pronunciation. We provide high-quality recitations from 30 distinct reciters at the verse level, representing diverse styles and dialectical nuances (see Table \ref{tab:dataset_stats} in the Appendix). Through careful validation and consistency checks, we ensure the reliability of the textual and audio components. By making this dataset available on Hugging Face, we aim to provide scholars, linguists, and AI researchers with a standardized resource that bridges textual, phonological, and semantic dimensions. This contribution supports the study of Qur’anic language and recitation traditions. It enables and promotes future work in cutting-edge computational applications at the intersection of speech, language, and Quranic studies.

%% file: section/2_RelatedWorks.tex
\section{Related Work}

There has been growing interest in developing resources for Qur’anic text and recitation, yet existing efforts typically focus on either textual analysis or audio recordings, rather than providing a fully integrated multimodal resource. Early structured corpora, such as the \emph{Quranic Arabic Corpus} \cite{dukhovny2009qac}, provide full text with detailed morphological and syntactic annotations, supporting parsing, part-of-speech tagging, and grammatical analysis. More recent contributions, including the \emph{MASAQ corpus} \cite{sawalha2024masaq} and the corpus introduced by \cite{nashir2025quranic}, expand these annotations and incorporate orthographic variants (Uthmani and Imlaai scripts), Buckwalter and phonetic transliterations, as well as English translations at the verse level. Similarly, the \emph{Tanzil project} \cite{tanzil2008} offers the Qur’an aligned with translations in more than 40 languages, supporting multilingual NLP research, though without detailed linguistic or recitational information.

\begin{table*}[t]
\centering
\caption{Feature comparison across prominent Quranic datasets. The proposed \textsc{Qur’an-MD} is the first to holistically integrate all listed modalities across multiple reciters at both verse and word levels, filling a critical gap in the available resources.}
\label{tab:comparison_final}
\resizebox{0.9\textwidth}{!}{%
\begin{tabular}{@{}lccccc@{}}
\toprule
 & \multicolumn{5}{c}{\textbf{Dataset}} \\
\cmidrule(l){2-6}
\textbf{Feature} & \begin{tabular}[c]{@{}c@{}}Quranic Arabic \\ Corpus \cite{dukhovny2009qac}\end{tabular} & \begin{tabular}[c]{@{}c@{}}Tanzil \\ Project \cite{tanzil2008}\end{tabular} & Tarteel \cite{khan2021tarteel} & QDAT \cite{qdat2021} & \begin{tabular}[c]{@{}c@{}}\textbf{\textsc{Qur’an-MD}} \\ \textbf{(Ours)}\end{tabular} \\ 
\midrule
Text (Arabic) & \yes & \yes & \yes & \yes & \yes \\
Morpho-syntactic Data & \yes & \no & \no & \no & \yes \\
Multi-Language Translation & \no & \yes & \no & \no & \yes \\
Word-level Audio & \no & \no & \no & \yes & \yes \\
Verse-level Audio & \no & \no & \yes & \yes & \yes \\
Multiple Reciters & \no & \no & \yes & \no & \yes \\ 
\bottomrule
\end{tabular}%
}
\end{table*}

Parallel to these textual resources, several initiatives have sought to capture Qur’anic recitation across multiple readers. Large-scale audio collections, such as \emph{Tarteel AI EveryAyah}\cite{khan2021tarteel} and the \emph{Quran-Recitations Dataset} \cite{quran_recitations_hf_dataset}, provide verse-level recordings across numerous reciters, while the \emph{Quranic Audio Dataset} \cite{quran_audio_dataset2024} and \emph{RetaSy} \cite{salameh2024retasy} focus on non-native reciters with labeled correctness annotations. The \emph{OpenSLR Quran Speech-to-Text corpus} \cite{openslr132} offers complete verse coverage for multiple reciters, and smaller targeted collections, such as \emph{QDAT} \cite{osman2021qdat} and its extensions \cite{eval_pronunciation_tajweed2025}, provide labeled data for Tajweed error detection using deep neural models. Recent Kaggle contributions, including the \emph{Quran Ayat Speech-to-Text} dataset \cite{bigguyubuntu2023ayat}, \emph{Quran Reciters dataset} \cite{omartariq2025}, \emph{Quran.com Audio dataset} \cite{abdo3id2022qurancom}, \emph{Quran Recitations for Audio Classification dataset} \cite{alrajeh2021recitations}, and the \emph{Comprehensive Quranic Dataset v1 (CQDV1)} \cite{cqdv12024}, offer large-scale verse-level audio recordings from dozens of reciters, but generally provide only Arabic text and audio without consistent transliteration or translation.

Taken together, these resources demonstrate substantial progress in both linguistic annotation and audio collection, yet they remain fragmented: either focusing on text or audio, verse- or word-level data. None combine verse and word-level text, transliteration, English translation, and audio across multiple reciters with consistent cross-modal alignment. \textsc{Qur’an-MD} addresses these gaps by integrating verse and word-level text, English translation, transliteration, word-level audio, and verse audio from 30 reciters, providing a unified multimodal resource that supports research across natural language processing, speech technologies, linguistic analysis, and digital Islamic studies.

%% file: section/3-4section.tex
\section{Dataset (\textsc{Qur’an-MD})}
\label{sec:dataset-preparation}

The dataset was constructed by aggregating and harmonizing three publicly available sources of Qur’anic data: (1) a Kaggle dataset of verse-level speech-to-text recordings by 30 reciters \cite{quran_ayat_speech_kaggle}, (2) the \texttt{quranwbw} repository containing aligned word-by-word Arabic text, English translations, and transliterations \cite{quranwbw_repo}, and (3) the Internet Archive collection of word-by-word Qur’an audio recordings \cite{quran_wordbyword_archive}. To unify these heterogeneous resources, we designed a hierarchical JSON template. At the top level, each \texttt{surah} (chapter) is represented by its numerical index (e.g., \texttt{"112"} for the 114 surah's) and annotated with its Arabic name, English translation, transliteration, and total verse count. Each surah contains a nested set of \texttt{verses} (ayahs), where each verse entry includes the Arabic text, its English translation, phonetic transliteration, paths to reciter-specific verse-level audio, and a list of word-level objects. Word objects store the word Arabic script, English translation, transliteration, and the corresponding aligned audio file. This structure allows seamless integration of textual, transliterated, and auditory modalities across both verse- and word-level granularities.  

The preparation pipeline proceeded in four main steps. First, surah-level metadata (surah number, names, and verse counts) was populated into the template. Second, from the Kaggle dataset, verse-level recitations by 30 reciters were aligned with their corresponding verse texts and added under the \texttt{audio\_ayah\_path} field. Third, word-level Arabic text, translations, and transliterations from the \texttt{quranwbw} repository were inserted into the template. Fourth, word-level audio files from the Internet Archive collection were matched to individual tokens and linked under the \texttt{audio\_word\_path} field. After populating all layers, automated validation scripts were applied to ensure correctness, checking that every word and verse was paired with its corresponding audio, and identifying and correcting missing or misaligned entries. This pipeline resulted in a complete, consistent, and multimodal representation of the Qur’an, ready for release as a standardized dataset.

\textbf{Dataset Format}: The dataset is organized in a hierarchical JSON structure, where each surah (chapter) of the Qur’an is represented as a nested object. This design ensures that verse-level and word-level information is consistently accessible and easily parsable for downstream applications. An example of Surah 112 (\textit{Al-Ikhlas}) is shown in Figure \ref{fig:example} (Appendix), highlighting metadata and audio/text linkage. This structure enables researchers to work seamlessly at both the verse level (for example, analyzing prosody between reciters) and the word level (for example, studying phoneme-grapheme alignment).

\begin{figure}[htp]
    \centering
\includegraphics[width=\textwidth]{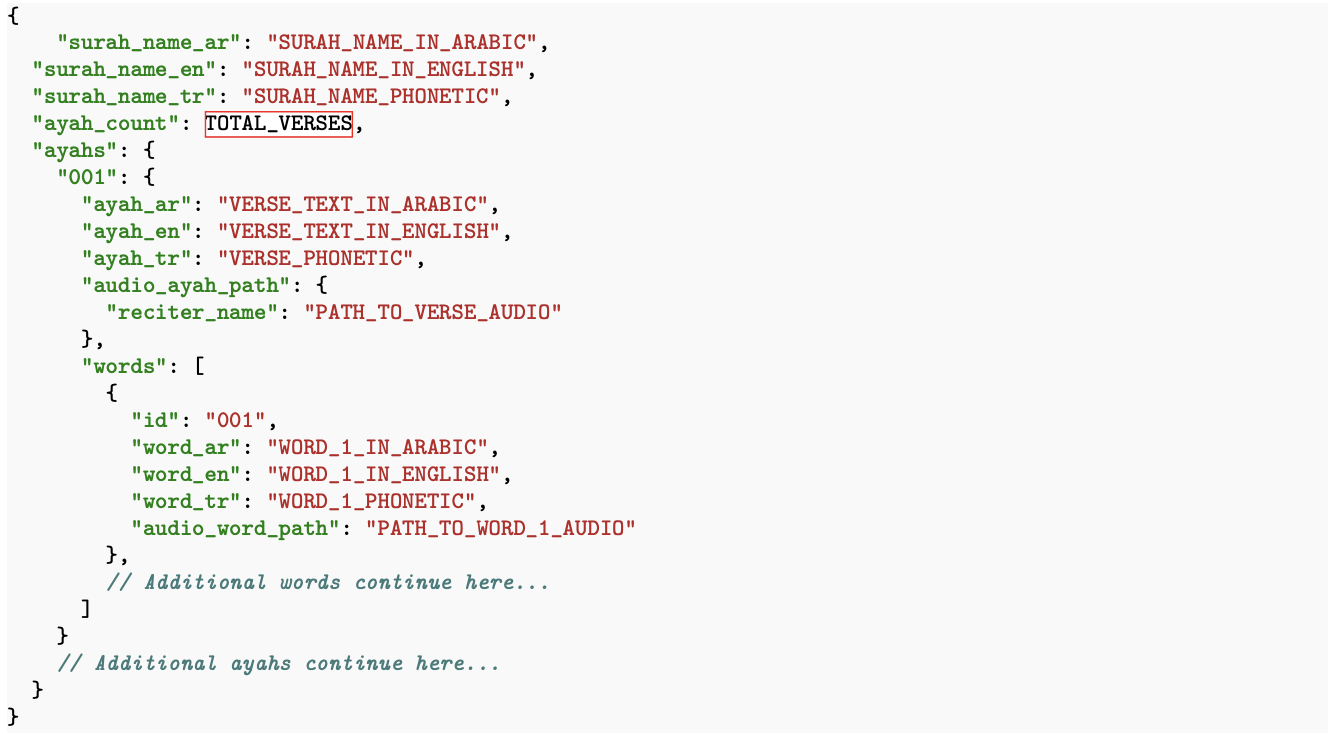} % Adjust width as needed
    % \caption{Example of format of Surah 112 (Al-Ikhlas) in the Dataset.}
    \label{fig:fing}
\end{figure}

%% file: section/5_FutureWorks.tex
\section{Future Work}

\textsc{Qur’an-MD} opens multiple avenues for future research at the intersection of speech processing, natural language processing, and digital Islamic studies. We outline three directions: (A) Speech-only tasks, (B) Multimodal approaches integrating speech and text, and (C) Tools and applications that can support both research and community engagement.  

\textbf{Speech-only Research}: A major direction lies in automatic speech recognition (ASR) for Qur’anic recitation, both at the verse and word level. The audio and verse-level transcripts aligned in the dataset with 30 reciters provide a unique testbed to develop end-to-end ASR models adapted to classical Arabic with the unique prosody and melodic contours in recitations. Evaluations can measure word error rate and segmentation accuracy across reciters, while challenges include modeling tajweed-driven prosody (madd, pauses, emphatics) and script-pronunciation mismatches.  

\textbf{Multimodal Speech-Text Research}: The full potential emerges when the modalities are combined. The dataset can drive progress in forced alignment and phoneme-level segmentation, producing high-quality timestamps for words and phonemes that benefit pronunciation modeling. It enables pronunciation tutoring systems tailored to Qur’anic recitation, which automatically detect tajweed issues and mispronounced words, providing corrective audio feedback and guidance on proper melodic and rhythmic patterns.

Another promising direction is Qur’anic TTS that respects tajweed rules and recitation styles. Leveraging parallel recordings from 32 reciters, the dataset enables the development of personalized TTS systems that provide corrective feedback for tajweed and Qur’anic melody in a user’s own voice, facilitating accurate imitation and learning. In addition, style transfer approaches can allow users to reproduce the prosody and melodic patterns of renowned reciters while preserving tajweed correctness, supporting both educational applications and expressive recitation synthesis. 

\textbf{Tools, Applications, and Community Resources:} \textsc{Qur’an-MD} also enables the development of advanced tools and applications that bridge research and community engagement. Multimodal embeddings of Qur’anic text and audio can support semantic retrieval of specific verses, style-aware search, and clustering of reciters, while facilitating the creation of robust tutoring systems for pronunciation and tajweed. Beyond machine learning applications, the dataset can be integrated into retrieval platforms, APIs, and interactive visualization tools for scholars, educators, and students, providing aligned verse- and word-level corpora that support linguistic, phonological, and prosodical research. Such resources can foster broader accessibility and engagement with the Qur'an and its studies.

% \textbf{Text-only / NLP Research}: On the text side, word-level annotations facilitate \emph{diacritization and morphological analysis} for classical Qur’anic Arabic, a long-standing challenge due to complex orthography. Incorporating aligned transliterations and pronunciations may further reduce diacritization errors. Another direction is \emph{semantic retrieval and question answering}, where embedded in aligned Arabic--English pairs can improve retrieval-augmented generation for Qur’anic queries. The dataset also enables research on \emph{alignment of translations and transliterations}, resulting in more accurate bilingual lexicons and improving cross-lingual applications in Islamic studies.  

%% file: section/6_Conslusion.tex
\section{Conclusion}
\label{sec:conclusion}

We curated \textsc{Qur’an-MD}, a comprehensive multimodal dataset of the Qur’an from three different sources that integrates textual and audio information with linguistic annotations at the verse- and word-level, encompassing 30 distinct reciters to capture the rich diversity of Qur’anic recitation. By aligning Arabic text, English translations, phonetic transliterations, and fine-grained audio recordings at both verse- and word-levels, the dataset enables a wide range of research applications, including automatic speech recognition, tajweed detection, pronunciation tutoring, personalized Qur’anic TTS, style transfer, and prosody modeling. Furthermore, it supports the development of multimodal embeddings for semantic retrieval, style-aware search, and reciter clustering, as well as interactive tools for scholars, educators, and learners. This resource bridges text and audio modalities at scale, providing a unique platform to advance computational modeling and applications in Qur’anic research.

%% file: section/7_appendix.tex
\newpage
\section{Appendix}  

Beyond ASR, the dataset can enable automatic tajweed detection and error classification, an area currently limited to small corpora like QDAT. In addition, it provides an opportunity to investigate the explainability of tajweed errors or mispronounced words, allowing models to highlight specific phonological or tajweed-related deviations and provide interpretable feedback for learners and researchers. Leveraging recitation at a word level, models could detect rules such as \emph{ghunnah, idgham, or madd}, with precision/recall measured against expert annotations. Similarly, the dataset supports reciter identification and style analysis, using embeddings to group readers by dialect or melodic tendencies, and prosody modeling, which quantifies melodic contours and pause structures for downstream use in Qur’anic TTS.

\begin{figure}[htp]
    \centering
\includegraphics[width=0.31\textwidth]{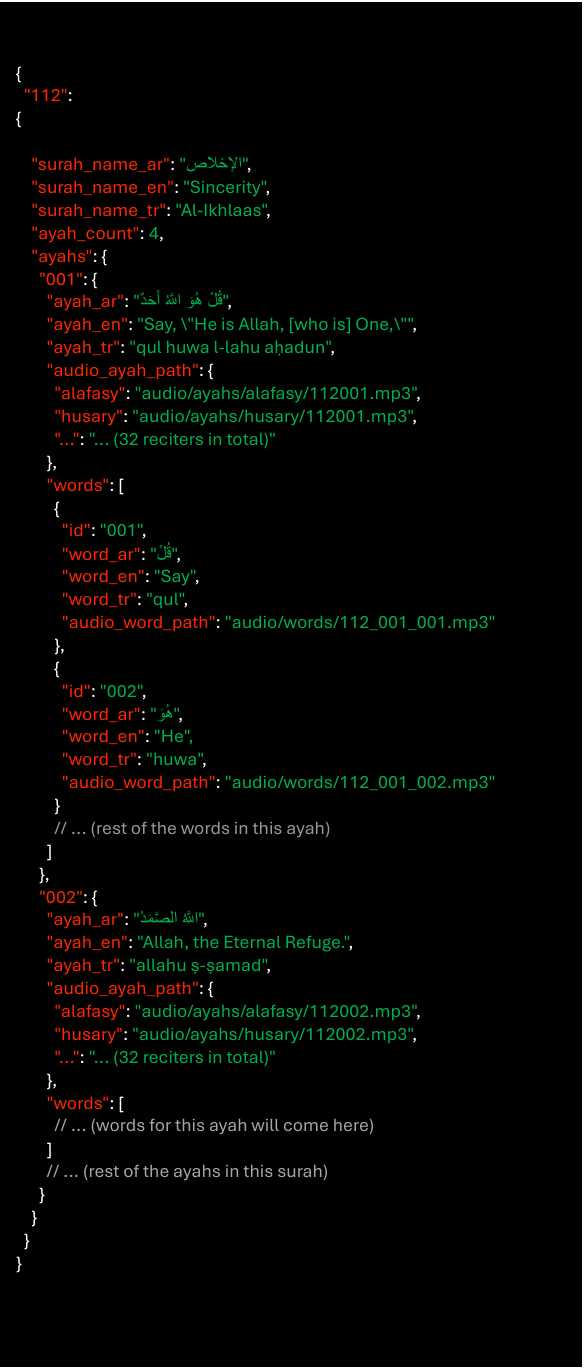} % Adjust width as needed
    \caption{Example of format of Surah 112 (Al-Ikhlas) in the Dataset.}
    \label{fig:example}
\end{figure}

\textbf{Example: Surah 112 (Al-Ikhlas)}: Figure \ref{fig:example} shows a simplified representation of the dataset format for the 112th surah (Al-Ikhlas). This example illustrates how each surah contains its metadata, how verses are indexed, and how both verse- and word-level audio/textual information are linked.

\begin{table}[ht]
\centering
\caption{Overview of the \textsc{Qur’an-MD} dataset.}
\label{tab:dataset_stats}
\begin{tabular}{@{}lll@{}}
\toprule
\textbf{Category} & \textbf{Attribute} & \textbf{Statistics / Details} \\
\midrule
\multirow{3}{*}{Corpus Size} 
  & Surahs & 114 \\
  & Ayahs & 6,236 \\
  & Words & $\sim$77.8k \\
\addlinespace
\multirow{3}{*}{Audio} 
  & Reciters & 30 (diverse styles) \\
  & Verse-level Audio & $\sim$665 hours \\
  & Word-level Audio & $\sim$22 hours \\
\addlinespace
\multirow{2}{*}{Modalities} 
  & Text & Arabic, English, Transliteration \\
  & Audio & Verse- and Word-level recordings \\
\bottomrule
\end{tabular}
\end{table}